\documentclass[conference]{IEEEtran}
\ifCLASSINFOpdf
\else
\fi
\usepackage{url}

\usepackage{graphicx}
\usepackage{verbatim}
\usepackage{caption}
\usepackage{subcaption}
\usepackage{multirow,datetime}
\usepackage{afterpage}
\usepackage{fixltx2e}

\newcommand\T{\rule{0pt}{2.6ex}}
\newcommand\B{\rule[-1.2ex]{0pt}{0pt}}

\begin{document}
%
\title{Expressing Ethnicity through\\ 
Behaviors of a Robot Character}

\author{\IEEEauthorblockN{Maxim Makatchev and Reid Simmons}
\IEEEauthorblockA{The Robotics Institute, Carnegie Mellon University\\
Pittsburgh, PA, USA\\
Email: \{mmakatch, reids\}@cs.cmu.edu}
\and
\IEEEauthorblockN{Majd Sakr and Micheline Ziadee}
\IEEEauthorblockA{Carnegie Mellon University in Qatar\\
Doha, Qatar\\
Email: msakr@qatar.cmu.edu, michelinekz@gmail.com\\
}}


%


\maketitle

\begin{abstract}

Achieving homophily, or association based on similarity, between a 
human user and a robot holds a promise of improved perception and task performance. 
However, no previous studies that address homophily via ethnic similarity with robots exist.
In this paper, we discuss the difficulties of evoking ethnic cues in a robot, 
as opposed to a virtual agent, and an approach to overcome those difficulties 
based on using ethnically salient behaviors. We outline our methodology for selecting and 
evaluating such behaviors, and culminate with a study that evaluates our hypotheses of the possibility
of ethnic attribution of a robot character through verbal and nonverbal behaviors and of
achieving the homophily effect.

\end{abstract}

\vskip 2mm


%
\IEEEpeerreviewmaketitle

\section{Introduction}

Individuals tend to associate disproportionally with others who are similar to themselves~\cite{Lazarsfeld1954}. This tendency, referred to by social scientists as \textit{homophily}, manifests itself with respect to similarities due to gender, race and ethnicity, social class background, and other sociodemographic, behavioral and intrapersonal characteristics~\cite{McPherson2001}. An effect of homophily has also been demonstrated to exist in human relationships with technology, in particular with conversational agents, where humans find ethnically congruent agents more persuasive and trustworthy (e.g.~\cite{NassIsbister2000,YinBickmore2010}). It is logical to hypothesize that homophily effects would also extend to human-robot interaction (HRI). However, to the best of our knowledge, no previous work addressing the hypothesis of ethnic homophily in HRI exists. 

One explanation for the lack of studies of homophily in HRI is the difficulty of endowing a robot with sociodemographic characteristics. While embodied conversational agents may be portraying human characters much like would be done in an animated film or a puppet theater, we contend that a robot, even with the state-of-the-art human likeness, retains a degree of its machine-like agency and its embeddedness in the real world (or, in theatric terms, a ``broken fourth wall''). Expressing ethnic cues with virtual agents is not completely problem-free either. For example, behaviors that cater to stereotypes of a certain community can be found offensive to the members of the community the agent is trying to depict~\cite{deRosisPelachaudPoggi2004}. We speculate that endowing non-humanlike robots with strong ethnic cues, such as appearance, may lead to similar undesirable effects.
Some ethnic cues, such as the choice of speaking American English versus Arabic, can be strong, but may not be very useful for a robot that interacts with multiple interlocutors in a multicultural setting. 

In this work, we argue for the need to explore more subtle, behavior-based cues of ethnicity. Some of these cues are known to be present even in interactions in a foreign language (see \cite{Aguilar1998} for an overview of pragmatic transfer). For example, it is known that politeness strategies, as well as conventions of thanking, apologizing, and refusing \textit{do} find their way into the second language of a language learner~\cite{Bardovi2007}. As we will show, even such subtle behaviors can be more powerful cues of ethnicity than appearance cues, such as a robot's face. 
We acknowledge the integral role of behaviors by referring to the combination of a robot's appearance and behaviors that aims to express a sociodemographic identity as a \textit{robot character}.


The term \textit{culture}, often used in related work, is ``one of the most widely (mis)used and
contentious concepts in the contemporary vocabulary'' \cite{agar2006}.
To avoid ambiguity, we will outline the communities of robot users that we are considering in terms of their \textit{ethnicity}, in particular, their native language, and, when possible, their country of residence. As pointed out in~\cite{Laitin2000}, even the concepts of a native language and mother tongue do not specify clear boundaries. Nevertheless, we will describe a multistage process where we (1)~attempt to identify verbal and nonverbal behaviors that are different between native speakers of American English and native speakers of Arabic, speaking English as a foreign language, then (2)~evaluate their salience as cues of ethnicity, and, finally, (3)~implement the behaviors on a robot prototype with the goals of evoking ethnic attribution and homophily. 



First, in Sections~\ref{sec:related} and \ref{sec:corpus} we outline various sources of identifying ethnically salient behavior candidates, including qualitative studies and corpora analyses. Then, in Section~\ref{sec:crowdsourcing}, we present our approach to evaluating salience of these behaviors as ethnic cues via crowdsourcing. In Section~\ref{sec:main}, we describe a controlled study with an actual robot prototype that evaluates our hypotheses of the possibility of evoking ethnic attribution and homophily via verbal and nonverbal behaviors implemented on a robot. We found support for the effect of behaviors on ethnic attribution but there was no strong evidence of ethnic homophily. We discuss limitations and possible reasons for the findings in Section~\ref{sec:discussion}, and conclude in Section~\ref{sec:conclusion}.








\section{Identifying ethnically salient behaviors}
\label{sec:identifying}
\subsection{Related work}
\label{sec:related}

Anthropology has been a traditional source of qualitative data on behaviors observed in particular social contexts, presented as ethnographies. Such ethnographies produce descriptions of the communities in terms of \textit{rich points}: the differences between the ethnographer's own expectations and what he observes~\cite{AgarBook1996}. For example, a university faculty member in the US may find it unusual the first time a foreign student addresses her as ``professor.'' The term of address would be a rich point between the professor's and the student's ways of using the language in the context. Note that this rich point can be a cue to the professor that the student is a foreigner, but may not be sufficient to further specify the student's ethnic identity.

Since ethnographies are based on observations of human interactions, their applicability in the context of interacting with a robot is untested. Worse yet, dependency on the identity of the researcher implies that a rich point may not be such to, for example, a researcher of a different ethnicity. The salience of behaviors as ethnic cues also varies from one observer to another. In spite of the sometimes contradictory conclusions of such studies, they provide important intuition about the space of candidates for culturally salient behaviors. 




Feghali, for example, summarizes the following linguistic features shared by native speakers of Arabic: repetition, indirectness, elaborateness, and affectiveness~\cite{Feghali1997}. Some of these linguistic features make their way into other languages spoken by native speakers of Arabic. Thus, the literature on pragmatic transfer from Arabic to English suggests that some degree of transfer occurs for conventional expressions of thanking, apologizing and refusing (e.g.~\cite{Bardovi2007} and \cite{Ghawi1993}). 

The importance of context has motivated several efforts to collect and analyze corpora of context-specific interactions. Iacobelli and Cassell, for example, coded the verbal and nonverbal behaviors, such as gaze, of African American children during spontaneous play~\cite{IacobelliCassell2007}. CUBE-G project collected a cross-cultural multimodal corpus of dyadic interactions~\cite{Rehm2009}.



In the next section, we describe our own collection and analysis of a cross-cultural corpus of receptionist encounters. The goal of our analysis is to identify behaviors that are potential ethnic cues of native speakers of American English and native speakers of Arabic speaking English as a second language, to these two ethnic groups. We will further refer to these ethnic groups as AmE and Ar, respectively.\footnote{We will also refer to the attribution of experimental stimuli, such as images of faces and robot characters as AmE and Ar. The exact formulation of the ethnic attribution item of questionnaires depends on the stimuli type.} Typically, when creating agents with ethnic identities (e.g.~\cite{IacobelliCassell2007}), the behaviors that are maximally distinctive (a kind of rich points, too) between the ethnic  groups are selected and implemented in the virtual characters. In this work, we propose an extra step of evaluating the salience of the candidate behaviors as ethnic cues via crowdsourcing. In addition to the argument that differentiates rich points from ethnic cues that we have given in the beginning of this section, such an evaluation allows one to pretest a wider range of behaviors via an inexpensive, online study, before committing to a final selection of ethnic cues for a more costly experiment with a physical robot colocated with the participants.

\subsection{Analysis of corpus of receptionist dialogues}
\label{sec:corpus}



\subsubsection{Participants and method}
We recruited AmE and Ar participants at Education City in Doha and at the CMU campus in Pittsburgh. Pairs of subjects were asked to engage in a role play where one would act as a receptionist and the other as a visitor. 
Before each interaction, the visitor was given a destination and asked to enquire about it in English.
Each pair of subjects had 2-3 interactions with one role assignment, and then 2-3 interactions with their roles reversed. 
The interactions were videotaped by plain view cameras. 

Among 2 Ar females, 5 Ar males, 4 AmE females and 1 AmE male acting as a receptionist, only one Ar male and one AmE female were considered as experts due to current or previous experience.

We annotated 46 of the interactions on multiple modalities of receptionist and visitor behaviors. 
Three of the interactions were annotated by a second annotator with the F-score of $0.88$ being the smallest of the F-scores of temporal interval annotations and transcribed word histograms.


%
%
%

\subsubsection{Results}



Due to the sparsity of the data, we limited quantitative analysis to gaze behaviors. Nonverbal behaviors, such as smiles and nods, and utterance content were analyzed as case studies.

Two gaze behaviors that took the longest fraction of interactions are \textit{gaze on visitor} and \textit{pointing gaze} (gaze towards a landmark while giving directions). 
The total amount of gaze on user grows linearly with duration and, contrary to prior reports on Arab nonverbal communication (see \cite{Feghali1997} for an overview), does not differ significantly between genders and ethnicities. 
Total durations of pointing gazes show more diversity. In particular, male receptionists pointed with their gaze significantly less than female ones.

Analysis of durations of continuous intervals of pointing gaze shows a trend towards short gazes (less than $1 s$) for Ar receptionists, compared to the prevalence of gazes lasting between 1 and $2 s$ for AmE receptionists. This trend is present for both female and male subjects. Correspondingly, Ar receptionists tend towards shorter (less than $1 s$) gaps between gazes on visitor. 


%
%


The observed wide range of pointing behaviors, including cross-cultural differences that are consistent across genders, motivated our selection of the frequency and duration of continuous pointing gaze as candidates for further evaluation.


We summarize our findings on realization of particular dialogue acts below.

\textbf{Greetings.}  Ar receptionists show a tendency to greet with ``hi, good morning (afternoon),'' and ``hi, how are you?''  AmE receptionists tended to use a simple ``hi,'' or ``hey,'' or ``hi, how can I help you?'' An expert AmE female receptionist used ``hi'' and an open smile with lifted eyebrows. An Ar male expert receptionist greeted an Ar male visitor with ``yes, sir'' and a nod in both of their encounters.

\textbf{Disagreement.} There are examples of an explicit disagreement with further correction (Ar female receptionist with Ar male visitor):
\begin{small}
\begin{verbatim}
v8r7: Okay, so I just go all the way down 
      there and...
r7v8: No, you take the elevator first.
\end{verbatim}
\end{small}

The expert Ar male receptionist, however, did not disagree explicitly with an Ar male visitor:
\begin{small}
\begin{verbatim}
r1v2: This way. Go straight.. (unclear) to 
      the end.
v2r1: So this (unclear) to the left.
r1v2: Yeah, to the right. Straight, to the 
      right.
\end{verbatim}
\end{small}

The same expert Ar male receptionist distanced himself from disagreeable information with another Ar male visitor:\footnote{Square brackets are aligned vertically to show overlapping speech, and colons denote an elongation of a sound.}
\begin{small}
\begin{verbatim}
v3r1: I think he moved upstairs, right? 
      Second floor. [He uses dean's office?]
r1v3:                [Mm::            ]  
      according to my directory it's one one 
      one zero zero seven.
\end{verbatim}
\end{small}

\textbf{Failure to give directions.} On 7 occasions the receptionist did not know some aspects of the directions to the destination. Some handled it by directing the visitor to a building directory poster or to search directions online. Others just admitted not knowing and did not suggest anything. Some receptionists displayed visible discomfort by not being able to give directions. On two occasions, receptionists offered an excuse for not knowing the directions:
\begin{small}
\begin{verbatim}
r6 (AmE female): I am not hundred percent 
    sure because I am new
r16 (Ar female): I don't know actually his name, 
    that's why I don't know where his office
\end{verbatim}
\end{small}

An AmE female (r15) used a lower lip stretcher (AU20 of FACS~\cite{Ekman1978}) and the emphatic: 

\begin{small}
\begin{verbatim}
r15v16: I have n.. I have absolutely no idea. 
        I've never heard of that professor and 
        I don't know where you would find it.
\end{verbatim}
\end{small}
She then went on to suggest that the visitor look up directions online. Interestingly, the visitor, an Ar female v16, who previously, as a receptionist, used an excuse and did not suggest a workaround, adopted r15's strategy and facial expression (AU20) when playing receptionist again with visitor v17:

\begin{small}
\begin{verbatim}
r16v17: I have no idea actually. you can look 
     up like online or something.
\end{verbatim}
\end{small}

The strategies, while not all specific to a particular native language condition, appear to be associated with expertise and perceived norms of professional behavior in the position of a receptionist. In the following section, we evaluate a subset of these behaviors for their ethnic salience. For more details on the analysis, we refer the reader to~\cite{makatchevThesis}.

\subsection{Crowdsourcing ethnically salient behavior candidates}

\label{sec:crowdsourcing}

Section~\ref{sec:related} highlighted the need to further evaluate behavior candidates suggested by analyses of corpora on their salience as ethnic cues. This section describes how we approach this task using crowdsourcing.


Crowdsourcing has been useful in evaluating HRI strategies, such as handling breakdowns \cite{MinkyungLee2010} and in cross-cultural perception studies, such as evaluating personality expressed through linguistic features of dialogues~\cite{Makatchev2011}. Here, we use crowdsourcing to estimate cross-cultural perception of verbal and nonverbal behaviors expressed by our robot prototype. 

\subsubsection{The robot prototype}

We used the Hala robot receptionist hardware~\cite{Simmons2011} to generate videos of the behavior stimuli for the crowdsourcing study and to serve as the robot prototype in the main lab experiment described in this paper (Fig.~\ref{fig:experiment}). The robot consists of a human-like stationary torso with an LCD mounted on a pan-tilt unit. The LCD ``head'' allows for relative ease in rendering both human-like and machine-like faces, as well as various appearance cues of ethnicity that we intend to control for in the following experiments.

\subsubsection{Influence of appearance and voice}

Ethnic salience of nonverbal behaviors is expected to be affected by the appearance and voice of the robot. Studies of virtual agents have shown that an agent's ethnic appearance may affect various perceptual and performance measures (see \cite{Baylor2004} for an overview).  To control for the appearance of the robot's face, we first conducted an Amazon's Mechanical Turk (MTurk) study that estimated ethnic salience of 18 images of realistic female faces that vary the tones of the skin,  hair and eye colors. 



\begin{figure*}[tbp]
  \begin{center}
  \begin{subfigure}[b]{0.2\textwidth}
    \centering
    \includegraphics[width=\textwidth]{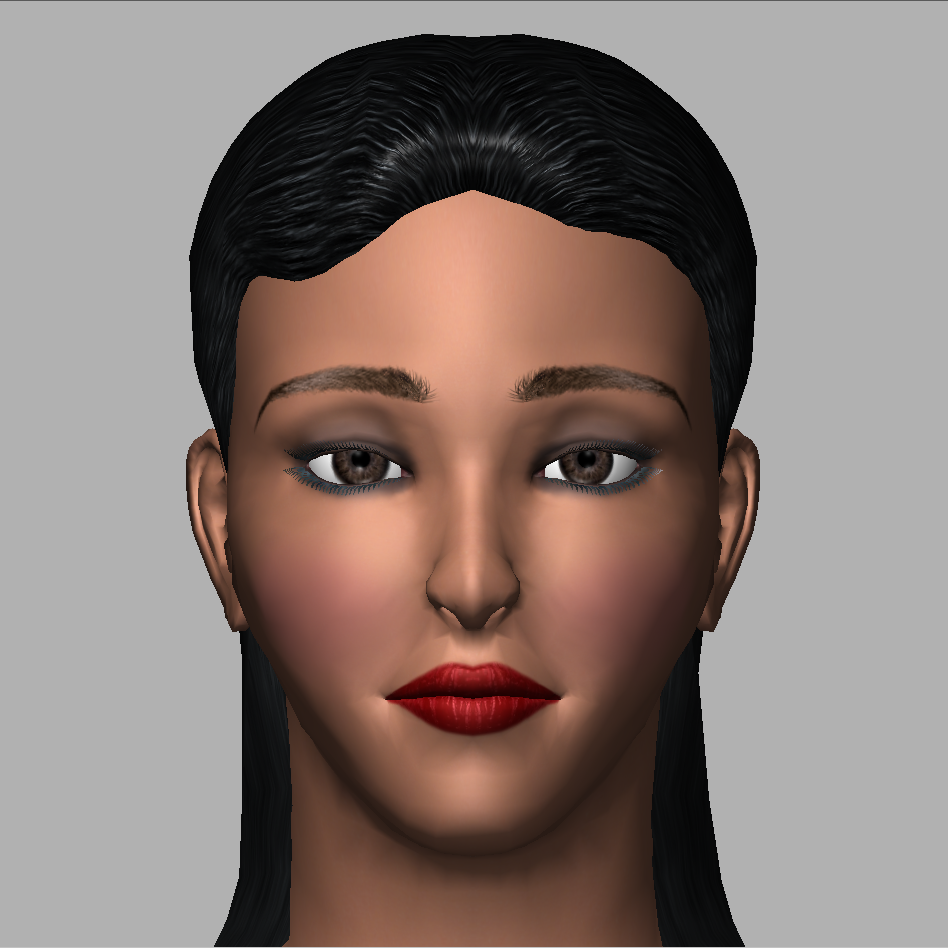}
    \caption{Face 1}
    \label{fig:head1}
  \end{subfigure}
~
  \begin{subfigure}[b]{0.2\textwidth}
    \centering
    \includegraphics[width=\textwidth]{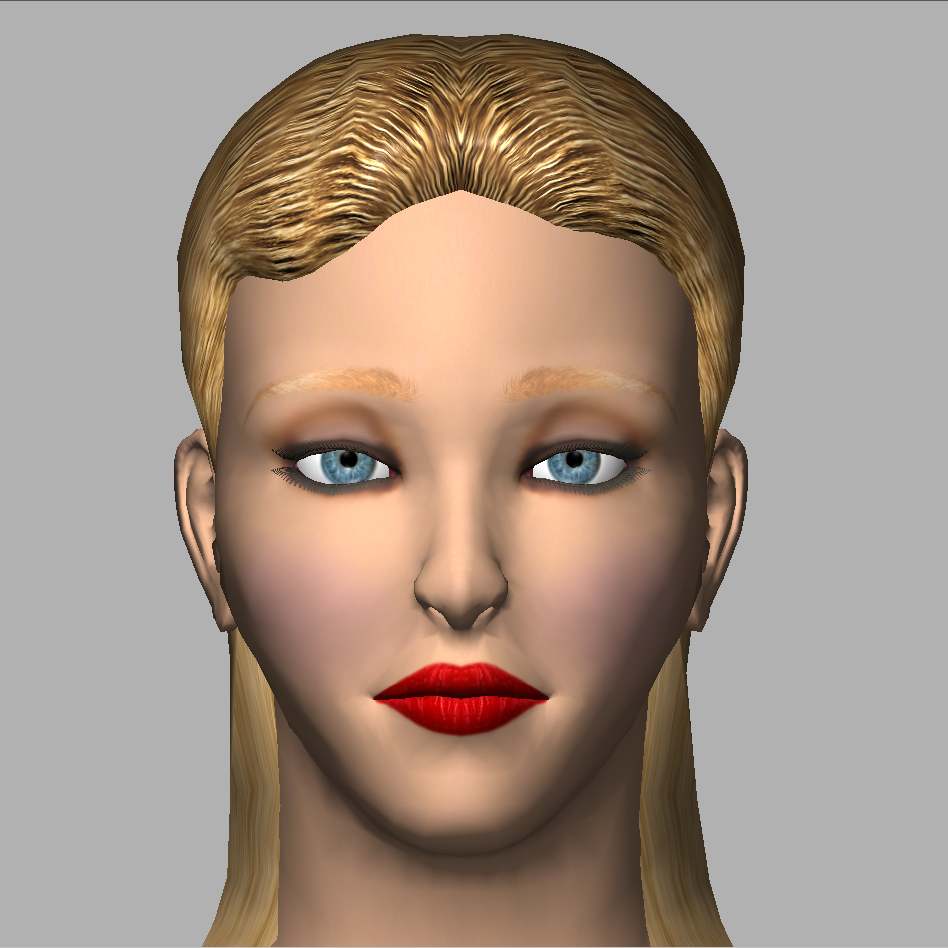}
    \caption{Face 2}
    \label{fig:head2}
  \end{subfigure}
~
  \begin{subfigure}[b]{0.2\textwidth}
    \centering
    \includegraphics[width=\textwidth]{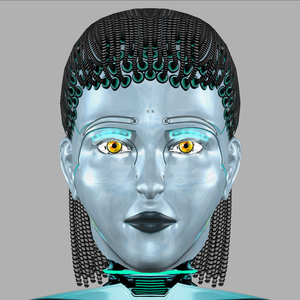}
    \caption{Face 3}
    \label{fig:head3}
  \end{subfigure}
~
  \begin{subfigure}[b]{0.2\textwidth}
    \centering
    \includegraphics[width=\textwidth]{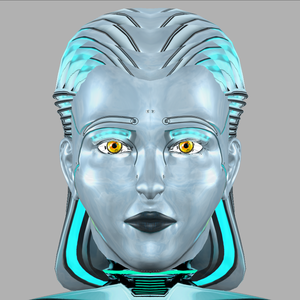}
    \caption{Face 4}
    \label{fig:head4}
  \end{subfigure}
  \caption{Faces 1 and 2 have high scores on their attribution as someone resembling native speakers of Arabic and American English, respectively. Faces 3 and 4 show low ethnic attributions.}
  \label{fig:heads}
\end{center}
\end{figure*}

In the end, we chose face 1 (Figure~\ref{fig:heads}), favored by both AmE and Ar subjects as someone resembling a speaker of Arabic, face 2, the face that was a clear favorite for Ar participants and that was among the highly rated by AmE participants as someone resembling a speaker of American English. Faces 1 and 2, together with two machine-like faces, face 3 and face 4, designed to suggest female gender and to minimize ethnic cues, were further evaluated on their ethnic attribution using two 5-point opposition scales (Not AmE---AmE, Not Ar---Ar). Ar participants gave high scores on attribution of face 1 as Ar, and face 2 as AmE, while giving low ethnic attribution scores to robotic faces 3 and 4. AmE participants, however, while attributing face 2 as clearly AmE, did not significantly differ their near neutral attributions of face 1, attributed face 3 as more likely Ar than AmE, and did not give high attribution scores to face 4. 


We also pretested 6 English female voices (available from Acapela Inc.) on their ethnic cues. 
Since voice cues are not in the scope of our study, we decided to use one of the voices that both AmE and Ar subjects scored as likely belonging to a speaker of American English.

\subsubsection{Measures and stimuli}

The stimuli categories of greetings, direction giving, handling a failure, disagreement, and politeness were presented to MTurk workers as videos sharing the web page with the Godspeed~\cite{Bartneck2008} questionnaire (with items measuring animacy, anthropomorphism, likeability, intelligence, and safety) and two questions on ethnic attribution. The ethnic attribution questions were worded as follows:
``The robot's utterances and movement copy the utterances and movements of a real person who was told to speak in English. Please rate the likely native language of that person, which is not necessarily the same as the language spoken in the video.'' The participants were asked to imagine that they were visiting a university building for the first time and the dialogue shown in the video happened between them and the receptionist, with the participant's utterances shown as subtitles.\footnote{\label{note:stretch}An example video can be seen at \url{http://youtu.be/6wA05CdFGUU} .} All stimuli were shown with faces 1, 2, and 3. For Ar participants, the task description and questionnaire items were shown in both English and Arabic. 

\textbf{Greetings.} Utterances ``yes, sir''/``yes, ma'am'' (depending on the reported gender of the MTurk worker) and ``hi'' were presented in combination with a physical robot head nod, an in-screen head nod, an open smile, or no movement besides mouth movements related to speaking. 
%
%
%
%

\textbf{Direction giving.} A fixed 4-step direction giving utterance was combined with 6 different gaze conditions: (1) moving the gaze towards the destination at the beginning of the first turn and moving the gaze back towards the user at the end of the fourth turn, (2) gazing towards the destination at every turn for $1.2 s$, (3) gazing towards the destination at every other turn for $1.2 s$, (4) gazing towards the destination at every turn for $0.8 s$, (5) gazing towards the destination at every other turn for $0.8 s$, (6) always looking towards the user (forward). Gaze switch included both the turning of the physical screen and graphical eye movement.

\textbf{Handling failure to provide an answer.} Dialogue 1 included emphatic admission of the lack of knowledge, lip stretcher (AU20) and a brief gaze to the left:\footnotemark[3]
\begin{small}
\begin{verbatim}
u: Can you tell me where the Dean's office is?
r: I have no absolutely idea.
\end{verbatim}
\end{small}

Dialogue 2 included an admission of the lack of knowledge, an excuse, and a brief gaze to the left:
\begin{small}
\begin{verbatim}
u: Can you tell me where the Dean's office is?
r: I don't know where it is, because I am new.
\end{verbatim}
\end{small}

\textbf{Handling a disagreement.} Two dialogues that do not involve any nonverbal expression and vary only on how explicit is the disagreement:
\begin{small}
\begin{verbatim}
u: Can you tell me where the cafeteria is?
r: Cafeteria... Go through the door 
   on your left, then turn right.
u: Go through the door, then left?
r: No, turn right.
\end{verbatim}
\end{small}

and

\begin{small}
\begin{verbatim}
u: Can you tell me where the cafeteria is?
r: Cafeteria... Go through the door 
   on your left, then turn right.
u: Go through the door, then left?
r: Yes, turn right.
\end{verbatim}
\end{small}

\textbf{Politeness}

The third direction-giving gaze condition (1.2$s$ gazes towards destination every other turn) was varied with respect to the politeness markers in the robot's utterances, such as ``please'' and ``you may'':
\begin{small}
\begin{verbatim}
u: Can you tell me where the library is?
r: Library... Go through the door on your left, 
   turn right. Go across the atrium, 
   turn left into the hallway and
   you will see the library doors. 
\end{verbatim}
\end{small}

and

\begin{small}
\begin{verbatim}
u: Can you tell me where the library is?
r: Library... Please go through the door on 
   your left, then turn right. You may go 
   across the atrium, then you may turn left 
   into the hallway and you will see the 
   library doors. 
\end{verbatim}
\end{small}

\subsubsection{Results}

Each of the stimuli categories above was scored during a within-subject experimental session by 16-35 MTurk workers in AmE and Ar language conditions.\footnote{We limit Ar participants to those with IP addresses from the following countries: Algeria, Bahrain, Egypt, Iraq, Jordan, Kuwait, Lebanon, Libya, Morocco, Oman, Palestinian Territories, Qatar, Saudi Arabia, Sudan, Syria, Tunisia, United Arab Emirates, Yemen.} We tested the significance of the behaviors as predictors of the questionnaire scores by fitting linear mixed effects models with the session id as a random effect and computing the highest posterior density (HPD) intervals. A selection of the significant results is presented below.

\textbf{Greetings.} Although AmE workers rated the verbal behavior ``yes, sir''/``yes, ma'am'' higher on likeability and intelligence, they gave it lower scores on AmE attribution. Surprisingly, the same verbal behavior scored higher on AmE attribution by Ar workers, but they did not consider it more likeable or intelligent. Smile and in-screen nod improved likeability for both AmE and Ar workers. However, AmE workers scored smile, physical nod, as well as face 3 as less safe, while Ar workers rated face 2 as less safe, and smile and in-screen nod as more safe nonverbal behaviors. 




\textbf{Direction giving.} 
AmE workers showed a main effect of gaze 6 on increasing AmE attribution (primarily due to interactions with face 1 and face 3) and on decreasing Ar attribution. Both Ar and AmE workers scored all faces with gaze 6 lower on animacy and likeability.  Face 2 had a main effect of lowering Ar attribution for Ar workers.

\textbf{Handling failure to provide an answer.}  Ar workers rated the no-excuse strategy as less Ar and as more AmE. AmE workers found face 1 with an excuse as less safe. Ar workers rated face 3 with no excuse as less likeable.

\textbf{Handling disagreement.} The explicit disagreement behaviors were ranked as more intelligent by AmE workers. Ar workers, on the other hand, scored face 3 giving an explicit disagreement as lower on likeability.

\textbf{Politeness.} Directions from face 3 without polite markers were scored as less likeable by  AmE and Ar workers. However, AmE rated polite directions from face 3 as less intelligent.

Averaging over faces, behaviors, stimuli categories, and participant populations, robot characters were attributed as more likely to be AmE rather than Ar ($M_{AmE}=3.90$, $M_{Ar}=1.70$, $t(5638.87)=28.07$, $p<0.001$). This overall prevalence of AmE attribution is present within each of the stimuli categories and within both AmE and Ar participants. It should not be surprising, as all characters spoke English, with the same voice. Interestingly, few combinations of behaviors and faces showed differences in perception between the two participant populations. Also, there were few combinations of behaviors and faces that resulted in any shift in ethnic attribution of the robot characters. For further details of this analysis we refer the reader to~\cite{makatchevThesis}.

Two main conclusions can be drawn. First, the ethnic attributions of the faces shown as images were generally not replicated by videos of the faces rendered on robots engaged in a dialogue. Second, since ethnic attributions rarely differ for these stimuli, other measures, such as Godspeed concepts, become more useful indicators of rich points. However, interpretation of the difference in Godspeed concept scores across subject groups is not that straightforward. For example, while AmE participants scored the robot with face~1, ``hi,'' and a smile higher only on likeability, Ar workers gave this combination higher scores also on intelligence, animacy, anthropomorphism, and safety. 
Is the corresponding behavior a good candidate for an ethnic cue? Even when ethnic attributions are significant, the subject groups may disagree on them. For example, the verbal behavior ``yes, sir''/``yes, ma'am'' has a negative effect on AmE attribution by AmE participants, but a positive effect on AmE attribution by Ar participants. This supports the idea that a behavior can be a cue of ethnicity $A$ when viewed by the members of ethnic group $B$ and not a cue when viewed by the members of $A$ itself. We decided on whether to consider such behaviors as ethnic cues by consulting with other sources, such as the corpus data and related work. For some behaviors, however, the results of this MTurk study were convincing on their own. For example, we have chosen gaze 6 (no pointing gaze) as one of the cues of AmE ethnicity for a followup study where the robot is colocated with the participants, described in the next section.


\section{Evaluating ethnic attribution and homophily}
\label{sec:main}

The analysis of the corpus of video dialogues and evaluation of behavior candidates via crowdsourcing allowed us to narrow down the set of rich points that are potential cues of ethnicity. We divided the most salient behaviors into two groups and presented them as two experimental conditions: behavioral cues of ethnicity Ar (BCE\textsubscript{Ar}) and behavioral cues of ethnicity AmE (BCE\textsubscript{AmE}). We hypothesize that these behavior conditions will affect perceived ethnicity of the robot characters.

\textbf{Ethnic attribution family of hypotheses 1Ar and 1AmE:} Behavioral cues of ethnicity will have a main effect on ethnic attribution of robot characters. In particular, BCE\textsubscript{Ar} will have a positive effect on the robot characters' attributions as Ar, and BCE\textsubscript{AmE} will positively affect the robot characters' attributions as AmE.

We also hypothesize that these behavior conditions will elicit homophily, measured as concepts of the Godspeed questionnaire and as an objective measure of task performance.

\textbf{Homophily family of hypotheses 2A--2F:} A match between the behavioral cues of ethnicity expressed by a robot character and the participant's ethnicity will have a positive effect on the perceived animacy (2A), anthropomorphism (2B), likeability (2C), intelligence (2D), and safety (2E) of robot characters. Participants in the matching condition will also show improved recall of the directions given by the robot character (2F).


\subsection{Participants}

Adult native speakers of Arabic who are also fluent in English and native speakers of American English were recruited in Education City, Doha. Data corresponding to one Ar subject and 3 AmE subjects were excluded from the analysis due to the less than native level of language proficiency or deviations in protocol. After that, there were 17 subjects (7 males and 10 females) in the AmE condition and 13 subjects (7 males and 6 females) in the Ar condition. The majority of the Ar participants were university students (mean age $19.5$, $SD=1.6$), while the majority of AmE participants were university staff or faculty (mean age $26.4$, $SD=9.2$).


\subsection{Procedure}

\begin{figure}[tbp]
  \begin{center}
    \includegraphics[width=0.38\textwidth]{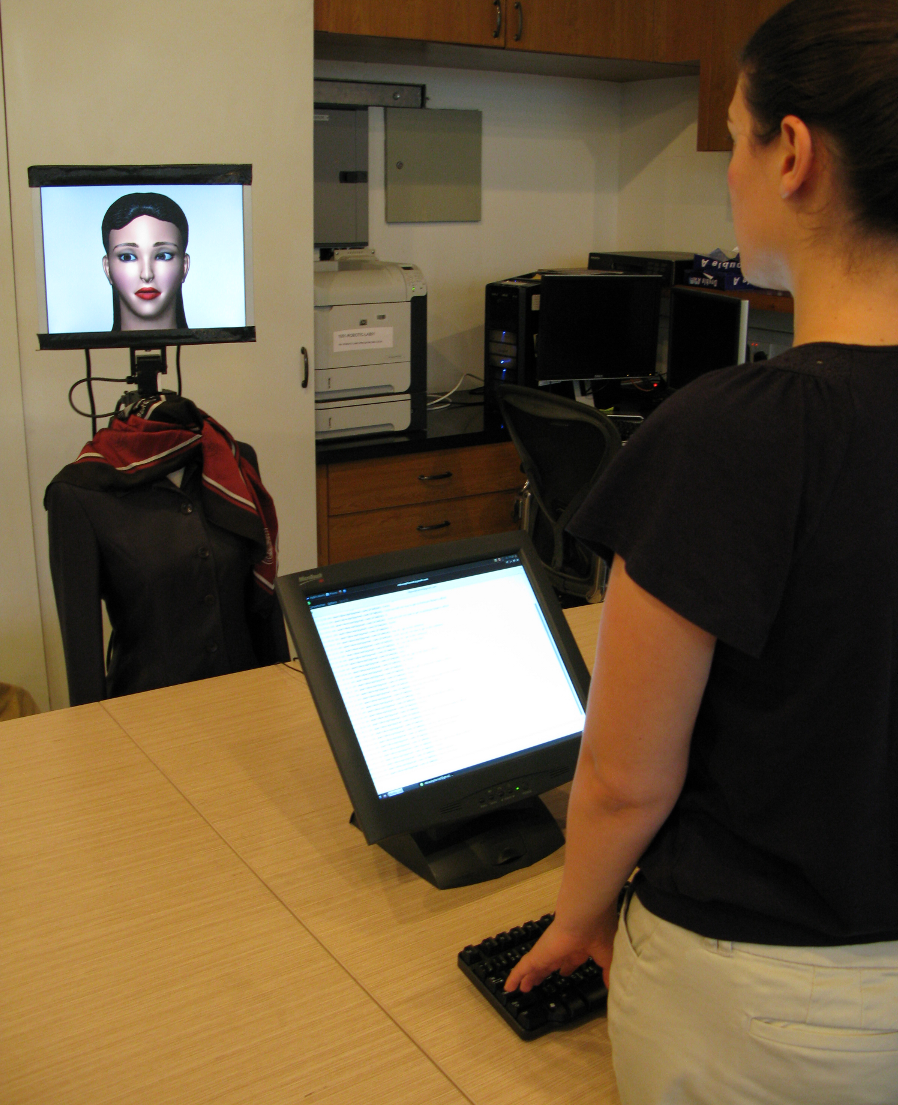}
    \caption{Experimental environment.}
    \label{fig:experiment}
\end{center}
\end{figure}

Experimental sessions involved one participant at a time. After participants completed a demographic questionnaire and the evaluation of their own emotional state (safety part of Godspeed questionnaire), they were introduced to the first robot character (Figure~\ref{fig:experiment}). Participants were instructed that they would interact with four different robot characters by typing in English, and were asked to pay close attention to the directions as they would be asked to recall them later. Participants were informed that, although the robot would reply only in English, it acts as a receptionist character of a certain ethnic background: either a native speaker of American English, or a native speaker of Arabic speaking English as a foreign language. Participants were asked to pay close attention to the robot's behaviors and the content of speech as they would be asked to score the likely ethnic background of the character played by the robot. 


Interaction with each robot character consisted of the three direction seeking tasks, with the order of the four characters for each of the participants varied at random. Behavior conditions BCE\textsubscript{Ar} and BCE\textsubscript{AmE} varied at random within the following constraints: for each of the participants, pairs face 1/face 2 and face 3/face 4 should be assigned to different behavior conditions. This allowed for within-subject comparison of behavior effect between two machine-like characters and of face effect between a machine-like and a realistic character. The robot's responses were chosen by an experimenter hidden from the user, to eliminate variability due to natural language processing issues. Post-study interviews indicated that participants did not suspect that the robot was not responding autonomously.

After the three direction-seeking tasks with a given character were completed, the participants were given a combination of the Godspeed questionnaire and two 5-point opposition scale questions addressing the likely native language of the character acted by the robot. All questionnaire items were presented in both Arabic and English.  Participants were also asked to recall the steps of the directions to the professor's office in writing and by drawing the route to the professor's office. 

\subsection{Stimuli}

The independent variables of the study were robot behavior (BCE\textsubscript{Ar} or BCE\textsubscript{AmE}), robot's face (1--4), and destinations and corresponding routes for the first dialogue (the professors' offices). The combinations of the conditions were mixed within and across-subjects.

Robot behaviors varied across dialogue acts of greeting, direction giving, handling disagreement and handling failure to provide directions. Combinations of verbal and nonverbal behaviors for each dialogue act and behavior condition are shown in Table~\ref{table:culture}. The robot's response to the user's final utterance (typically a variation of ``bye'' or ``thank you'') did not vary across conditions and was chosen between ``bye'' or ``you are welcome,''\footnote{The response to thanks was, erroneously, ``welcome'' for the first 2 AmE and 6 Ar subjects. We controlled for this change in the analysis.} as appropriate.

\begin{table}[htb]
\begin{center}
\begin{tabular}{|p{1.5cm}|p{3.1cm}|p{3.1cm}|}
\hline
Dialogue act\T \B & BCE\textsubscript{Ar} & BCE\textsubscript{AmE} \\ \hline\hline
Greetings \T &  ``Yes sir (ma'am)'' +\newline virtual nod & ``Hi'' +\newline open smile \B \\ \hline
Directions  \T & $0.8 s$ pointing gaze  at each step + politeness \B & Constant gaze on user \\ \hline
Disagreement \T &  \T No explicit contradiction: ``Yes, turn right'' \B \T &  \T An explicit contradiction: ``No, turn right''  \T \\ \hline
Handling failure  \T & \T Admission of failure +\newline brief gaze away +\newline providing an excuse & Emphatic admission of failure + brief gaze away\newline + lower lip pull \B \T \\ \hline

\end{tabular}
\end{center}
\caption{The manipulated verbal and nonverbal behaviors.}
\label{table:culture}
\end{table}

The directions to the professor's offices started from the actual experiment site and corresponded to a map of an imaginary building. Each route consisted of 6 steps (e.g.~``turn left into the hallway'') with varying combinations of turns and landmarks.  The landmarks were used in the template ``walk down the hallway until you see a [landmark], then turn [left or right],'' once per route. Since we use the recall of directions as a measure of task performance, the directions were pretested to be feasible, but not trivial to memorize.

Each participant conducted 3 direction-seeking tasks with each of the four faces (12 tasks overall). In the first task, participants were asked to greet the robot, ask directions to a particular professor's office, and end the conversation as they deemed appropriate. In the second task, the participants had to ask the robot for directions to the cafeteria, and, after the robot gives directions, which are always ``cafeteria... go through the door on your left, then turn right,'' to imagine that they misunderstood the robot and ask to clarify if they should turn left after the door. The robot would then respond with the disagreement utterance, corresponding to conditions BCE\textsubscript{Ar} or BCE\textsubscript{AmE}. In the third task, the participants had to ask the robot for directions to the Dean's office, for which the robot would execute one of the failure handling behaviors (see Table~\ref{table:culture}).

\subsection{Results}

Scores on the items of the Godspeed questionnaire and two items of ethnic attribution were used as perceptual measures of the robot. We use the success or failure of locating the sought professor's office on the map as a binary measure of task performance. 
Godspeed scales showed internal consistency with Cronbach's alpha scores above 0.7 for both AmE and Ar participants. All the $p$-values are given before correction for 2-hypothesis familywise error for hypotheses 1Ar and 1AmE, or 6-hypotheses familywise error for hypotheses 2A--2F. 

\textbf{Attribution hypothesis 1Ar.}
The data does not indicate a main effect of BCE or faces on Ar attribution. However, a stepwise backward model selection by AIC suggests a negative effect of the interaction between face 4 and BCE\textsubscript{AmE} on the attribution ($F[3,100]=3.44$, $p=0.020$). The significance of the interaction terms between faces and BCE is confirmed by comparing linear mixed effects models with subjects as random effects using the likelihood ratio test: $\chi^2(3, n=120) = 9.48$, $p=0.024$. The 95\% highest posterior density (HPD) intervals for the coefficient for the interaction between BCE\textsubscript{AmE} and face 2 is $[-2.14, -0.01]$ and for the coefficient for the interaction between BCE\textsubscript{AmE} and face 4 is $[-2.07, -0.04]$.

\begin{figure*}[tbp]
 \begin{center}
  \begin{subfigure}[b]{0.8\textwidth}
    \centering
    \includegraphics[width=\linewidth]{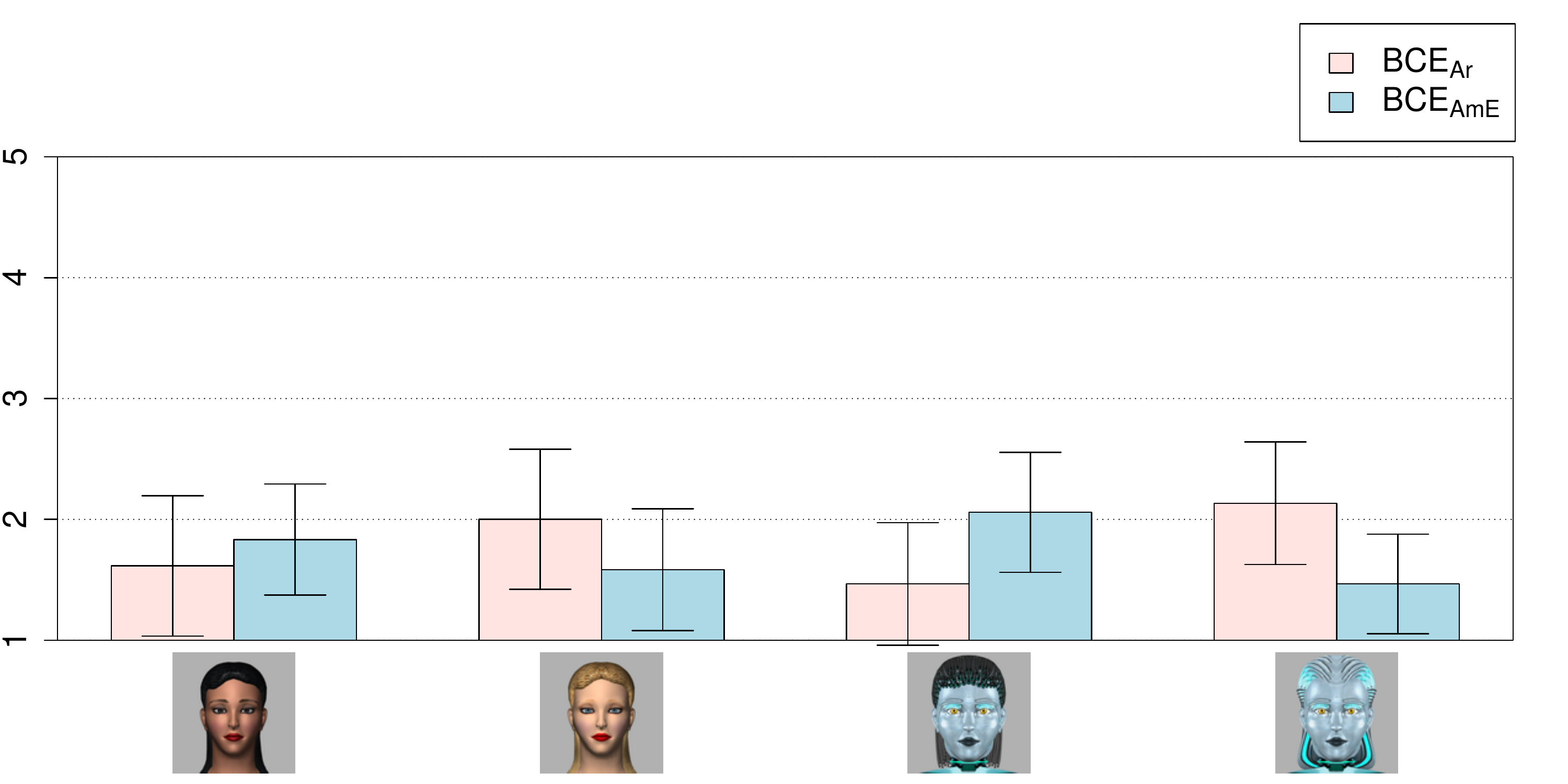}
    \caption{Female and male participants.}
    \label{fig:ar_face_bce}
  \end{subfigure}
\\
  \begin{subfigure}[b]{\textwidth}
    \centering
    \includegraphics[width=0.8\linewidth]{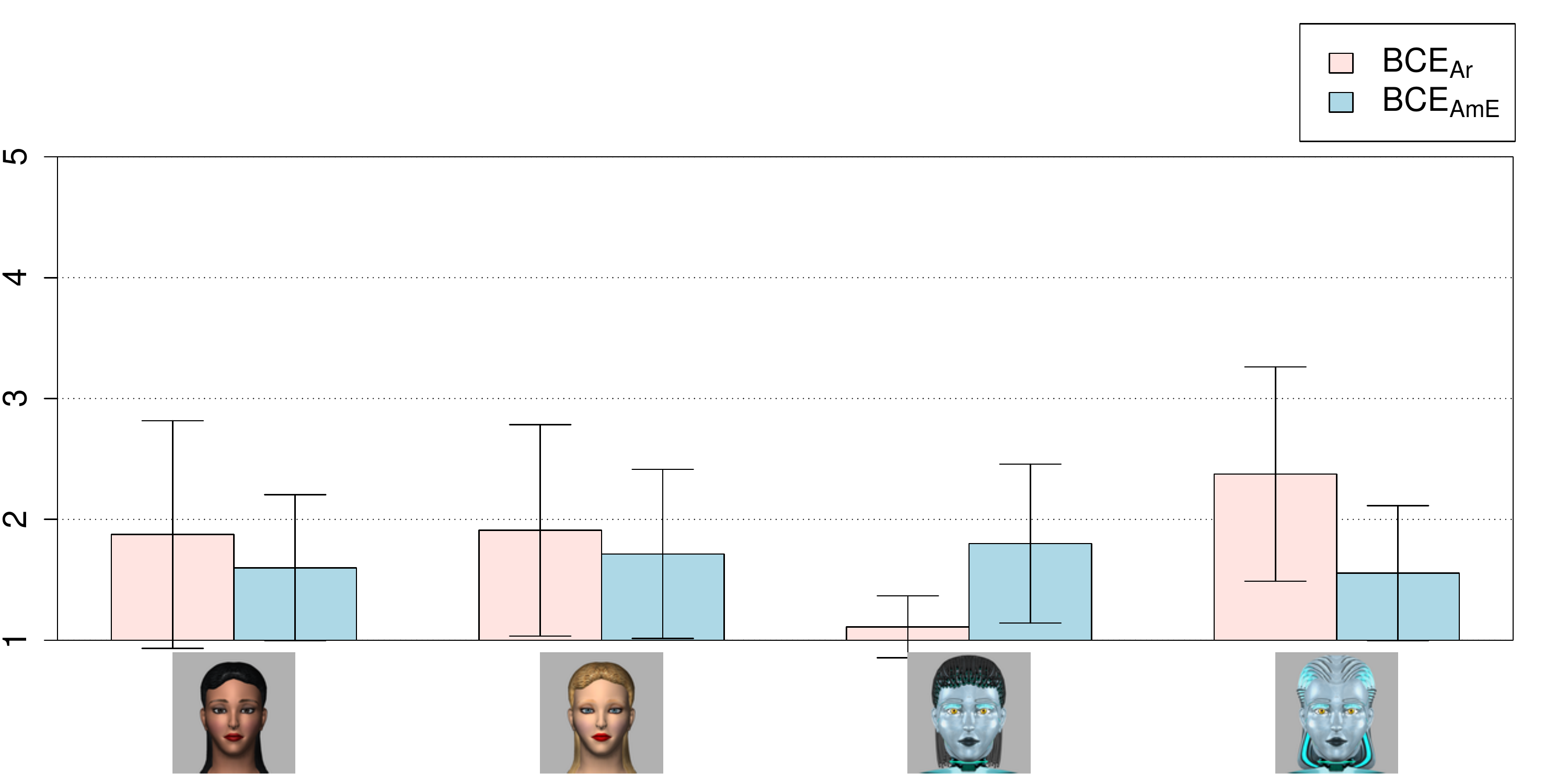}
    \caption{Female participants only.}
    \label{fig:ar_face_bce_f}   
  \end{subfigure}
\\
  \begin{subfigure}[b]{\textwidth}
    \centering
    \includegraphics[width=0.8\linewidth]{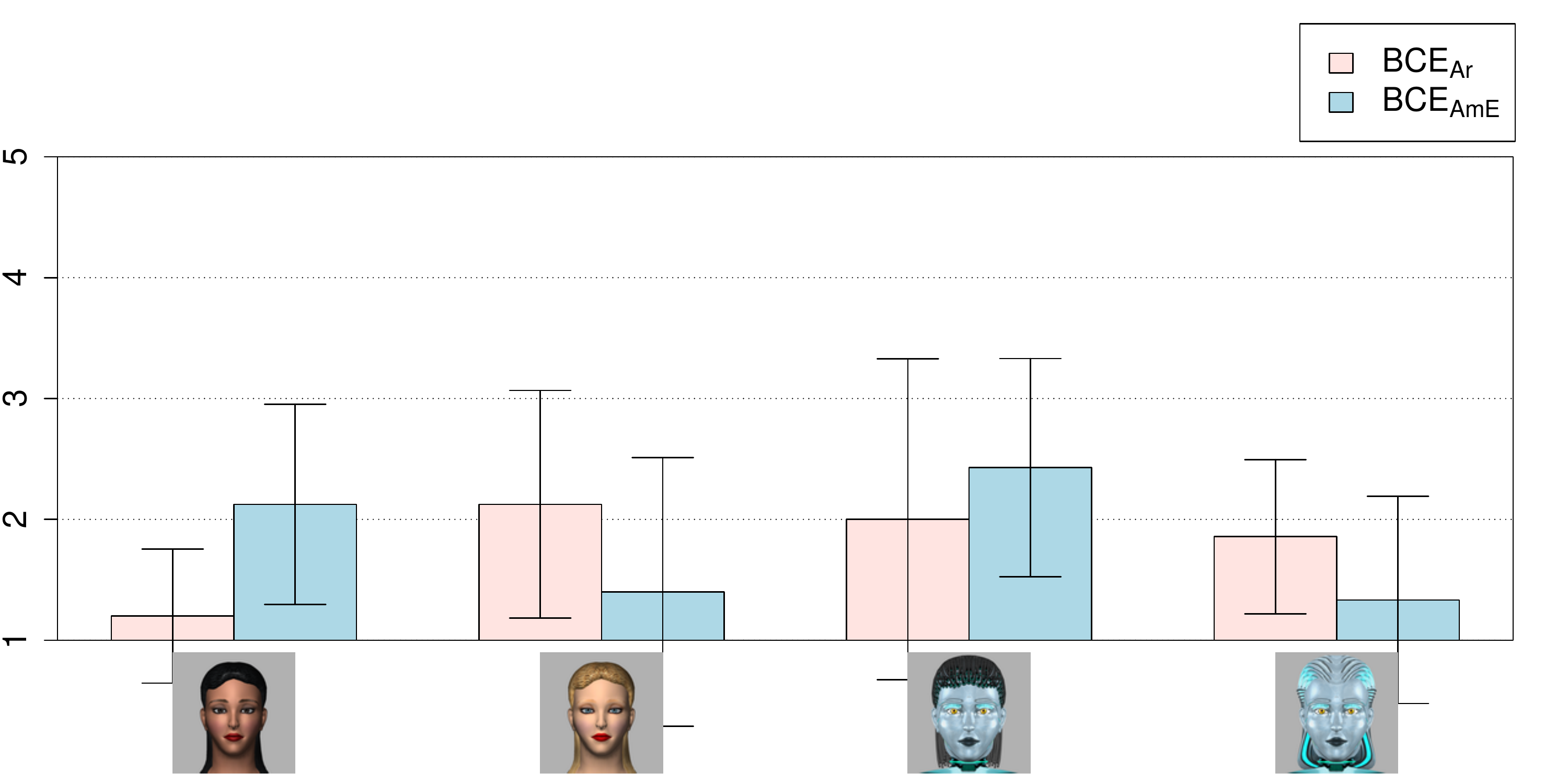}
    \caption{Male participants only.}
    \label{fig:ar_face_bce_m} 
  \end{subfigure}
\end{center}
    \caption{Score means on attribution of the robot characters as a native speaker of Arabic. Brackets correspond to 95\% confidence intervals. These plots are for visualization only, as direct pairwise comparison would not account for subject effects.}
    \label{fig:ar_face_bce3}
\end{figure*}

While there is no evidence of a main effect of the faces, the results suggest that with respect to the attribution of the robot characters as native speakers of Arabic the participants were sensitive to manipulation of BCE only for the robot characters with faces 2 and 4 (Figure~\ref{fig:ar_face_bce}). Indeed, focusing on the conditions with face 2 and 4 yields the effect of the BCE term with $\chi^2(1, n=59) = 7.89$, $p=0.005$ and the entirely negative 95\% HPD interval for BCE\textsubscript{AmE} $[-1.19, -0.15]$. Focusing on BCE\textsubscript{AmE} conditions, on the other hand, yields the effect of faces with $\chi^2(3, n=60) = 9.60$, $p=0.022$, but all 95\% HPD intervals for the faces trap 0.

\textbf{Attribution hypothesis 1AmE.} The likelihood ratio test for the mixed effects models supports the main effect of BCE on the robot characters' attributions as AmE, $\chi^2(1, n=120) = 5.44$, $p=0.020$. Mean scores of the robot's attribution as AmE are $M_{BCE\textsubscript{Ar}}=3.62$ ($SD=1.21$) and $M_{BCE\textsubscript{AmE}}=4.00$ ($SD=0.92$). The significance of the interaction between BCE and gender is $\chi^2(1, n=120) = 4.14$, $p=0.042$. Controlling for an additive face term, the significance of BCE is $\chi^2(1, n=120)=4.88$, $p=0.027$ and the significance of BCE and gender interaction is $\chi^2(1, n=120) = 5.02$, $p=0.025$. The HPD intervals for the model's coefficients that do not trap 0 are $[0.20, 1.15]$ for BCE\textsubscript{AmE} and $[-1.41, -0.02]$ for the interaction term between BCE\textsubscript{AmE} and males. 

\begin{figure*}[tbp]
 \begin{center}
  \begin{subfigure}[b]{0.8\textwidth}
    \centering
    \includegraphics[width=\linewidth]{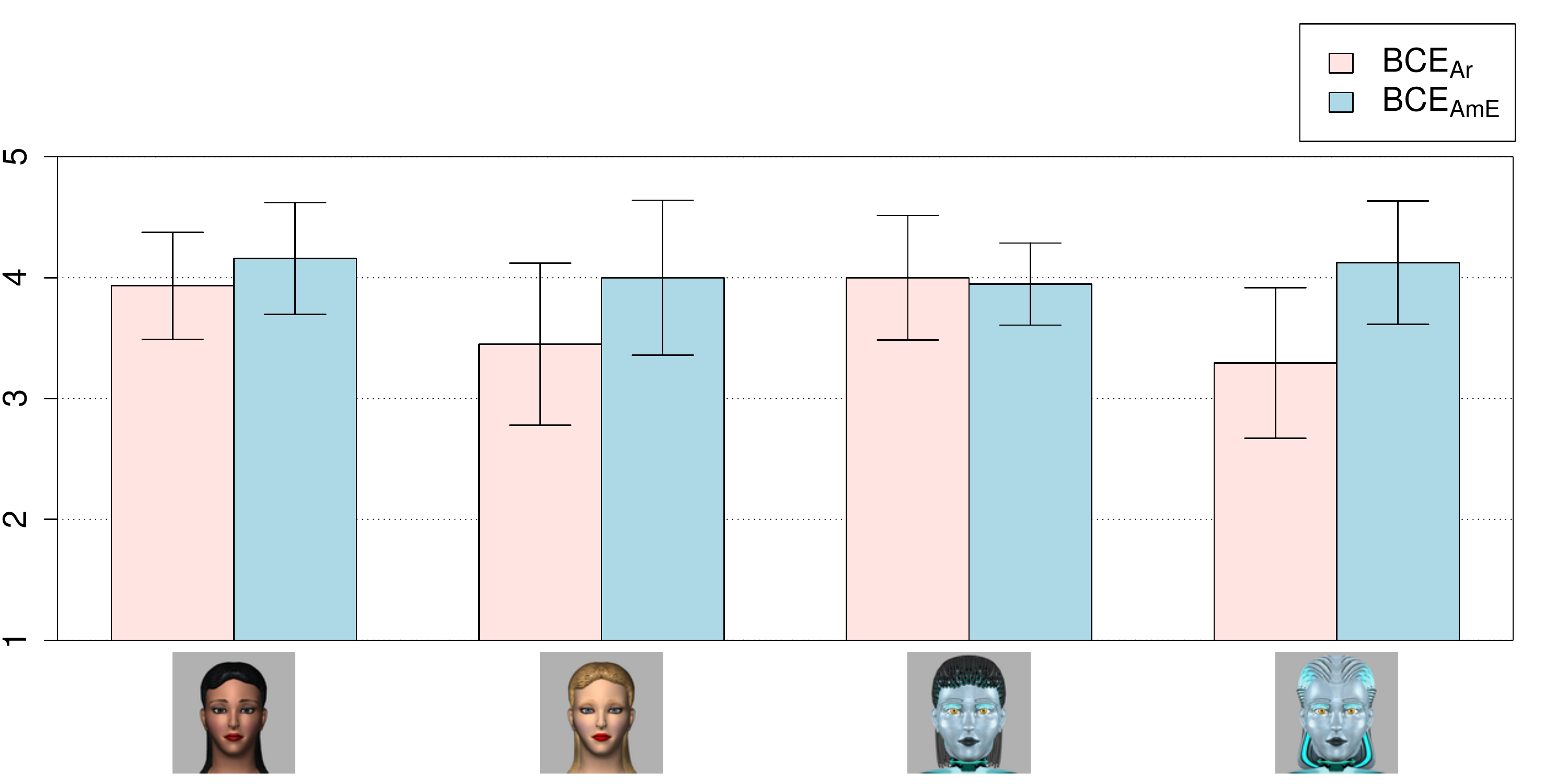}
    \caption{Female and male participants.}
    \label{fig:enus_face_bce}
  \end{subfigure}
\\
  \begin{subfigure}[b]{\textwidth}
    \centering
    \includegraphics[width=0.8\linewidth]{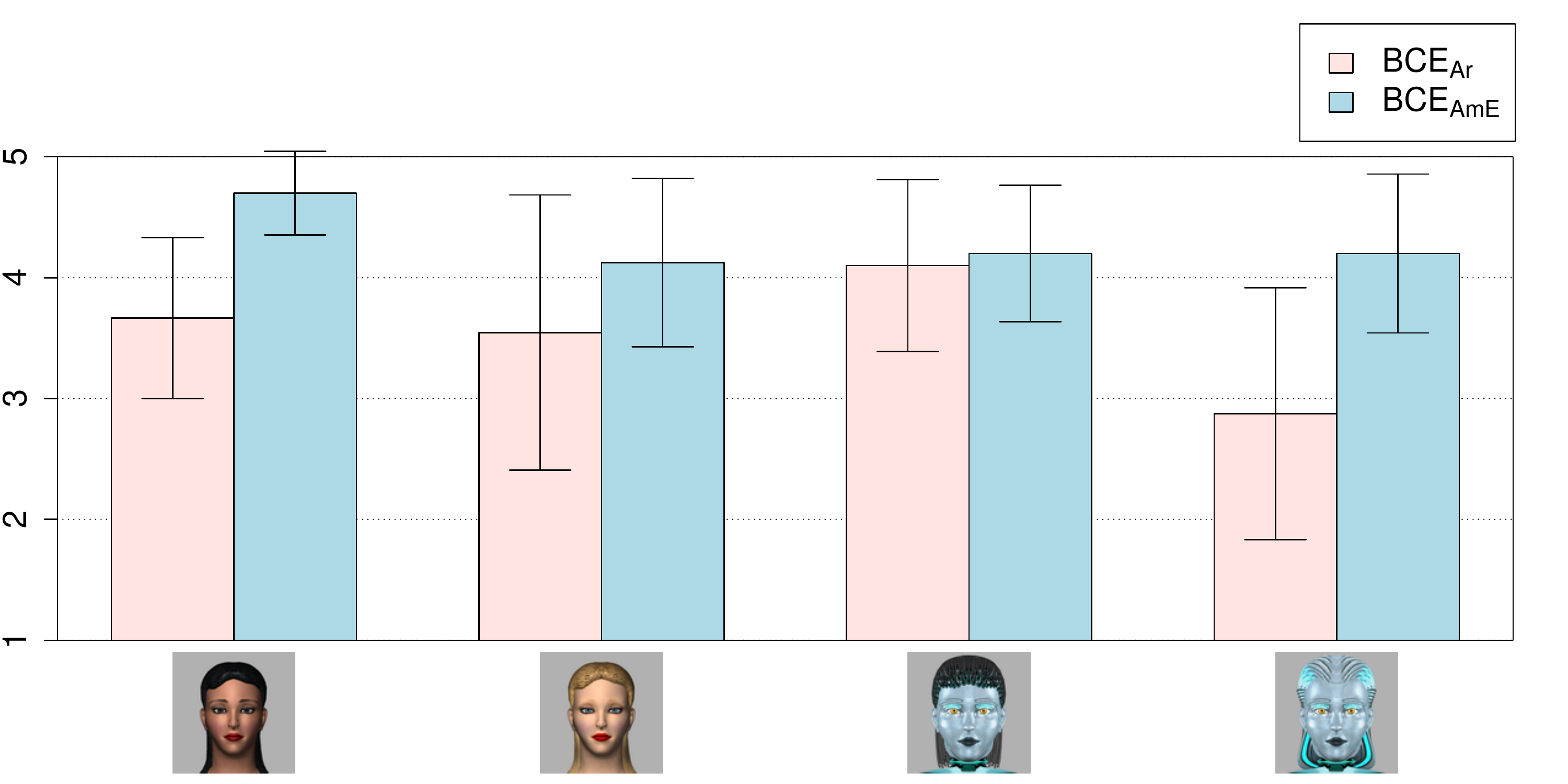}
    \caption{Female participants only.}
    \label{fig:enus_face_bce_f}   
  \end{subfigure}
\\
  \begin{subfigure}[b]{\textwidth}
    \centering
    \includegraphics[width=0.8\linewidth]{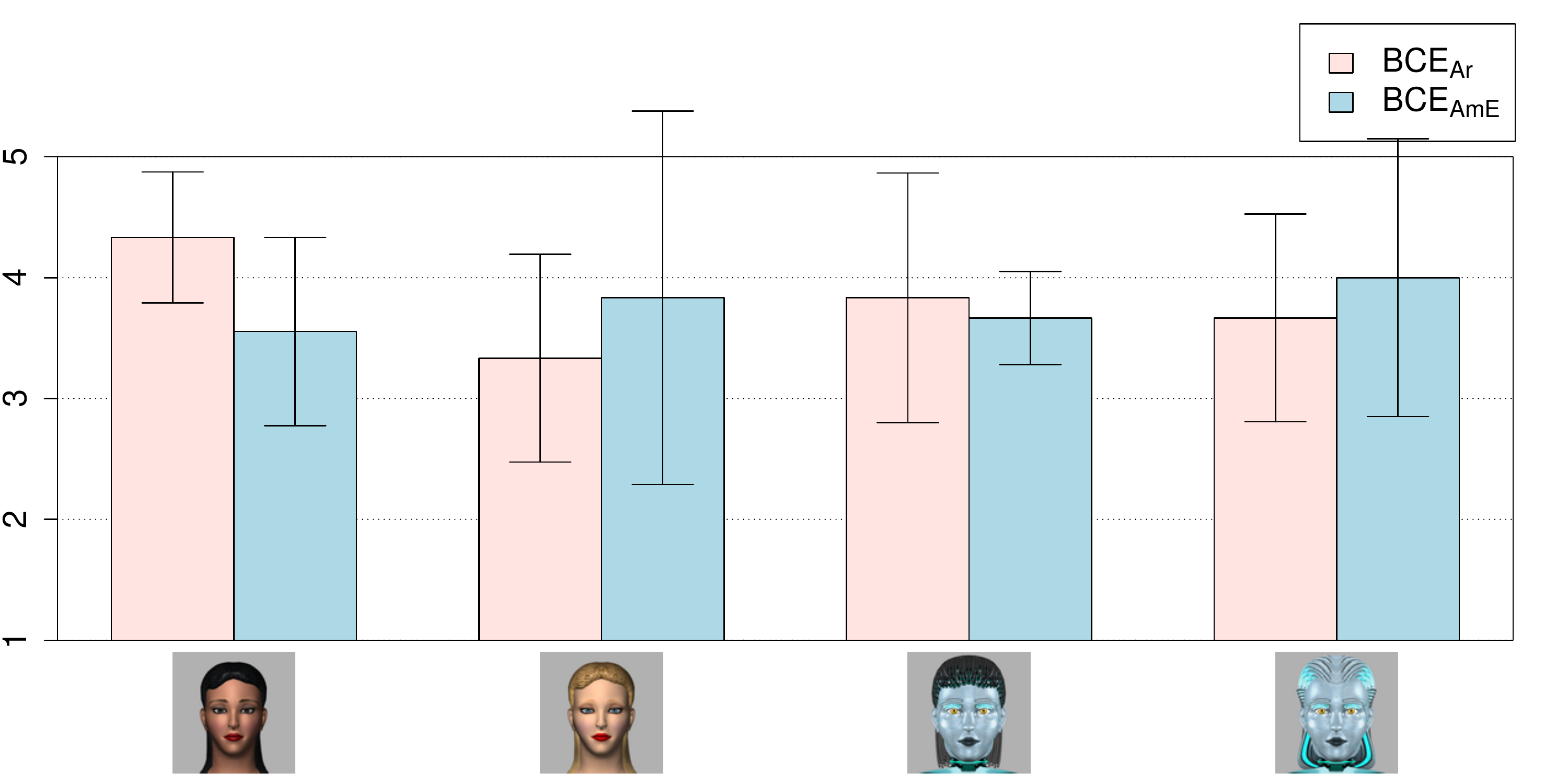}
    \caption{Male participants only.}
    \label{fig:enus_face_bce_m} 
  \end{subfigure}
\end{center}
    \caption{Score means on attribution of the robot characters as a native speaker of American English. Brackets correspond to 95\% confidence intervals. These plots are for visualization only, as direct pairwise comparison would not account for subject effects.}
    \label{fig:enus_face_bce3}
\end{figure*}

The negative effect of interaction between males and BCE\textsubscript{AmE} suggests that the behavioral cues of ethnicity would have a stronger effect among female subjects. Indeed, testing for the significance of BCE for female subjects yields $\chi^2(1, n=64)=8.74$, $p=0.003$. The effect of BCE on male subjects does not suggest significance (Figure~\ref{fig:enus_face_bce3}).


The analysis supports the hypothesis that varying behavioral cues of ethnicity from BCE\textsubscript{Ar} to BCE\textsubscript{AmE} has a positive effect on the attribution of the robot characters as native speakers of American English. This effect is more evident among females. There is no significant evidence of a main effect of the faces.

\textbf{Homophily hypotheses 2A--2F.} Tests of the interaction effects between the behavioral cues of ethnicity and the participant's native language do not support any of the homophily hypotheses. We present below a selection of interesting results showed by further exploratory analysis.

BCE\textsubscript{AmE} characters were rated as more animate, $\chi^2(1, n=120) = 6.00$, $p=0.014$. A combination of BCE\textsubscript{AmE} and face 4 was significantly more likeable, $t(119)=2.27$, $p=0.024$, in particular for AmE participants. Relatively to other participants, Ar males gave the robot characters lower safety scores, $t(119)=4.48$, $p<0.0001$.

Task performance, in terms of success or failure of locating the professor's office on the map, did not show any significant associations with the independent variables. For further details, we refer the reader to~\cite{makatchevThesis}.

\section{Discussion}
\label{sec:discussion}

The effect of behaviors on the perceived ethnicity of the robot characters was more evident in the attributions as a native speaker of American English rather than as a native speaker of Arabic. 
The fact that the behaviors did have an effect on attribution of the robot characters as Ar with faces 2 and 4 suggests that the manipulated behaviors were not just strong enough to distinguish between native and non-native speakers of American English, but also had relevance to the robot's attribution as Ar. The colocation of the questionnaire items for AmE and Ar attributions, however, could have suggested to the participants that these two possibilities are mutually exclusive and exhaustive. The correlation between the attribution scores is moderate, albeit significant, $r=-0.44$, $p<0.0001$. An improved study design could evaluate only one of the attributions per condition or per participant.

While behaviors may be stronger cues of ethnicity than faces for robot characters, there is an evidence supporting a potential of complex interactions between the two. One explanation for the lack of the main effect of faces is a potentially insufficient degree of human likeness of our robot prototype. Post-study interviews indicated that many participants had difficulties in interpreting the ethnic attribution questions in the context of the robot. It would be interesting to replicate the study with more anthropomorphic robots that could, potentially, more readily allow attributions of ethnicity, such as Geminoid~\cite{Ishiguro2005}.



Other limitations of this study include the fixed voice, with a strong attribution as AmE by participants of both ethnic groups, and the typed entry of user utterances. Some of the participants missed the robot's nonverbal cues as they were focusing their gaze on the keyboard even after they finished typing. We selected the typed, rather than spoken user input, out of the concern that participants would not buy into the idea of the robot's autonomy, especially since some of them are familiar with the prototype (although with a different face) as a campus robot receptionist that relies on typed user input. This concern may be less relevant for uninitiated participants, in which case an improved Wizard-of-Oz setup could rely on spoken user input.

Only AmE participants' data, only for some of the faces, was consistent with a possibility of behavior-associated ethnic homophily. The possible reasons for the almost complete lack of any homophily effects in our data include (a)~the highly international environment, with most of the participants being students or faculty of Doha's American universities, and (b)~a potentially low sensitivity of Godspeed items and the chosen objective measure to the hypothesized homophily effects. It is also possible that the effect on ethnic attribution, while significant, was not large enough to trigger ethnic homophily. There is, however, evidence that in human encounters ethnically congruent behaviors can trigger homophily even when the ethnicities do not match (e.g.~\cite{Dew1993} and literature on cultural competence).

\section{Conclusion}
\label{sec:conclusion}

Achieving ethnic homophily between humans and robots has the lure of improving a robot's perception and user's task performance. This, however, had not previously been tested, in part due to the difficulties of endowing a robot with ethnicity. We tackled this task by attempting to avoid overly obvious and potentially offensive labels of ethnicity and culture such as clothing, accent, or ethnic appearance (although we control for the latter), and instead by aiming at evoking ethnicity via verbal and nonverbal behaviors. We have also emphasized the robot as a performer, acting as a character of a receptionist. Our experiment with a robot of a relatively low human likeness shows that we can evoke associations between the robot's behaviors and its attributed ethnicity. Although we did not find evidence of ethnic homophily, we believe that suggested pathway can be used to create robot characters with higher degree of perceived similarity, and better chances of evoking homophily effect.
  

The methodology of selecting candidate behaviors from qualitative studies and corpora analyses, evaluating their salience via crowdsourcing, and finally implementing the most salient behaviors on a physical robot prototype is not limited to ethnicity, but can potentially be extended to endowing robots with other aspects of human social identities. In such cases, crowdsourcing does not only help to alleviate the hardware bottleneck inherent in HRI but can also facilitate recruitment of study participants of the target social identity.




\section*{Acknowledgment}
This publication was made possible by NPRP grant 09-1113-1-171 from the Qatar National Research Fund (a member of Qatar Foundation). The statements made herein are solely the responsibility of the authors. 
The authors would like to thank Michael Agar, Greg Armstrong, Brett Browning, Justine Cassell, Rich Colburn, Fr\'ed\'eric Delaunay, Imran Fanaswala, Carol Miller, Suhail Rehman, Michele de la Reza, Candace Sidner, Lori Sipes, Mark Thompson, and the reviewers for their contributions to various aspects of this work.



\bibliographystyle{IEEEtran}
%


\end{document}